\documentclass{article}
\usepackage{arxiv}
\usepackage[utf8]{inputenc}
\usepackage[T1]{fontenc}
\usepackage{cite}
\usepackage{amsmath,amssymb,amsfonts}
\usepackage{graphicx}
\usepackage{algorithm,algorithmic}
\usepackage{hyperref}
\usepackage{url}
\usepackage{booktabs}
\usepackage{microtype}
\hypersetup{hidelinks}
\usepackage{textcomp}
\usepackage[table]{xcolor}
\usepackage{wrapfig}
\def\BibTeX{{\rm B\kern-.05em{\sc i\kern-.025em b}\kern-.08em
    T\kern-.1667em\lower.7ex\hbox{E}\kern-.125emX}}

\newcommand{\mat}[1]{\mathbf{#1}}
\newcommand{\MAP}{\text{MAP}}
\newcommand{\DSA}{\text{DSA}}
\newcommand{\DRR}{\text{DRR}}
\newcommand{\nat}{\text{nat}}
\newcommand{\can}{\text{can}}

\begin{document}
\title{GeoPose: Patient-agnostic CTA-to-DSA registration through projection-space calibration}

\author{
 Rudolf L. M. van Herten\thanks{Corresponding author} \\
  Weill Cornell Medicine, Cornell Tech\\
  New York, NY 10044 \\
  \texttt{rlv4001@med.cornell.edu} \\
   \And
 Robert Graf \\
  Technical University of Munich\\
  81675 Munich \\
  \texttt{robert.graf@tum.de} \\
  \And
 Paula Feldman \\
  Weill Cornell Medicine, Cornell Tech\\
  New York, NY 10044 \\
  \texttt{paf4004@med.cornell.edu} \\
  \And
 Johannes C. Paetzold \\
  Weill Cornell Medicine, Cornell Tech\\
  New York, NY 10044 \\
  \texttt{jpaetzold@med.cornell.edu} \\
}

\maketitle

\begin{abstract}
Aligning intraoperative biplanar digital subtraction angiography (DSA) to pre-procedural computed tomography angiography (CTA) requires rapid and accurate 3D-to-2D registration. Optimization-based methods are sensitive to initialization and may require hundreds of iterations, whereas learning-based approaches commonly rely on patient-specific training. We propose GeoPose, a population-trained framework that estimates the C-arm pose in a learned canonical frame and transfers it to the native frame of an unseen CTA through projection-space calibration and transform composition. A population-trained residual network refines the pose, followed optionally by low-budget image-driven optimization. GeoPose requires neither patient-specific adaptation nor explicit inter-volume preregistration. On 80 DSA observations from 20 held-out patients, optimization-free GeoPose achieved a carotid mean projected centerline distance (mPCD) of 5.8 mm and a clDice of 0.45, compared with 14.5 mm and 0.28 for the best-performing baseline, while requiring only 0.15 s. After 25 optimization iterations, GeoPose reached an mPCD of 4.6 mm and a clDice of 0.58 in approximately two seconds. Under the same budget, native-initialized optimization achieved 14.6 mm and 0.15, respectively. GeoPose thus provides rapid native-frame registration with fixed population-level weights and the geometric correspondence required for downstream biplanar 3D vascular reconstruction. Code is available on \href{https://github.com/RoelvH97/GeoPose}{GitHub}.
\end{abstract}

\keywords{Computed tomography angiography \and domain generalization \and digital subtraction angiography \and image registration \and pose estimation}
\section{Introduction}
\label{sec:introduction}
Acute ischemic stroke is a neurological emergency in which treatment delay translates directly into irreversible cerebral tissue loss~\cite{powers2020acute}. This is particularly consequential for large-vessel occlusions, which impact a large downstream vascular network that perfuses cerebrovascular tissue. Endovascular thrombectomy (EVT) substantially improves functional outcomes in this setting~\cite{berkhemer2015randomized}, but its benefit declines rapidly with delayed reperfusion~\cite{nogueira2026time}. Consequently, any computational assistance introduced into the clinical workflow must provide actionable information within a strict time window and without adding patient-specific preparation that could delay treatment.

EVT follows pre-procedural CT angiography (CTA), which provides three-dimensional anatomical and vascular context, and is directly aided by intraoperative biplanar digital subtraction angiography (DSA)~\cite{powers2019guidelines}. DSA serves as the gold standard for intraoperative guidance for its ability to capture high-resolution vascular dynamics~\cite{fonseca2021european,yang2026topbrain,shaban2022digital}. Each DSA image is a two-dimensional projection with no direct spatial correspondence to the CTA. Estimating the posteroanterior (PA) and lateral (LAT) C-arm poses would therefore establish correspondence, allowing CTA-derived 3D segmentations and treatment targets to be projected onto the acquired DSA views. Furthermore, the recovered poses may provide a geometric basis for 3D vascular reconstruction from biplanar images.

\begin{figure*}
     \includegraphics[trim=0cm 0cm 0cm 0cm, width=1.\textwidth]{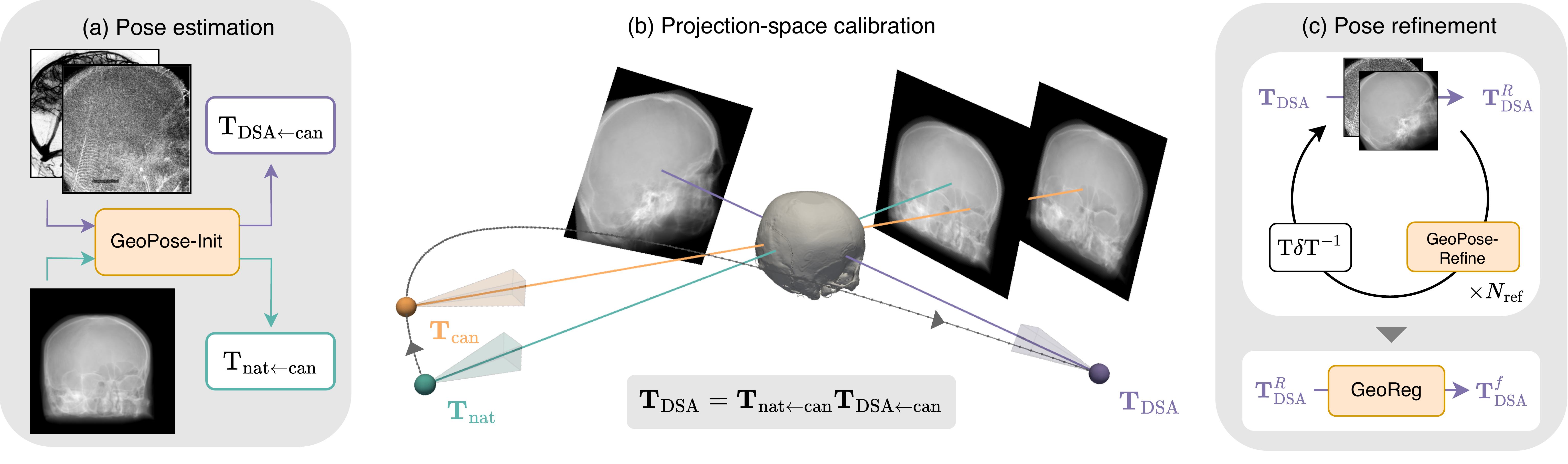}
     \caption{Overview of the proposed CTA-to-DSA registration pipeline. \textit{(a) Pose estimation}: GeoPose-Init regresses a canonical-frame transform from a DSA maximum intensity projection ($\mathbf{T}_{\text{DSA}\leftarrow\text{can}}$) and a calibration transform from a canonical isopose DRR of the patient CTA ($\mathbf{T}_{\text{nat}\leftarrow\text{can}}$). \textit{(b)~Projection-space calibration}: their composition maps the prediction into the patient's native volume frame, absorbing template misalignment and providing an initial pose estimate~$\mathbf{T}{_\text{DSA}}$. \textit{(c) Pose refinement}: GeoPose-Refine applies up to $N_{\text{ref}}$ image similarity-gated residual updates $\mathbf{T}{\delta\mathbf{T}}^{-1}$ from re-rendered DRRs to obtain the refined pose $\mathbf{T}_{\text{DSA}}^{R}$, which is subsequently test-time optimized by GeoReg to the final pose $\mathbf{T}_{\text{DSA}}^{f}$. The recovered native-frame pose establishes spatial correspondence between the pre-procedural CTA and intraoperative DSA, enabling CTA-based anatomical roadmapping in the acquisition views and providing the geometry required for downstream biplanar 3D vascular reconstruction.}
     \label{fig:main}
\end{figure*}

Most 3D-to-2D registration methods estimate these poses by comparing observed X-rays with digitally reconstructed radiographs (DRRs) rendered from the CTA volume. Differentiable X-ray rendering has made it possible to optimize this alignment directly using gradient descent-based methods~\cite{gopalakrishnan2022fast,gopalakrishnan2024intraoperative}. Nevertheless, the optimization objective is highly non-convex: registration remains sensitive to initialization, and recovering from a broad pose distribution may require hundreds of rendering and optimization steps~\cite{unberath2021impact}. Learning-based pose estimators can improve the capture range by amortizing pose initialization. For example, xvr combines population-level pretraining with patient-specific adaptation, reducing the required fine-tuning to approximately five minutes~\cite{gopalakrishnan2025rapid}. LXPose instead replaces test-time refinement with a cascaded pose-regression network and achieves real-time inference, at the cost of requiring a separate model for every target volume~\cite{facente2026toward}. Although effective in planned interventions, patient-specific network training is undesirable in acute stroke, where even short additional processing stages compete with an already time-critical treatment pathway.

GeoReg addresses a complementary limitation of cerebrovascular registration by aligning DSA and CTA without requiring vascular segmentation~\cite{van2026georeg}. GeoReg derives temporal maximum-intensity projection (MAP) X-ray silhouettes directly from DSA sequences and jointly optimizes their alignment with CTA-derived DRRs under a soft biplanar geometric constraint. This formulation avoids the need to extract corresponding vascular structures across modalities, but its practical speed and robustness remain governed by the initial pose. A default pose initialization may require hundreds of test-time optimization steps and does not account for the coordinate-frame differences between the learned population and a new patient's native, unregistered CTA.

To address these limitations, we introduce \emph{GeoPose}, a population-trained framework for rapid CTA to biplanar DSA registration. GeoPose estimates the C-arm pose of each DSA-derived MAP in a shared canonical frame, where template alignment gives pose a consistent meaning across patients by learning a pose--content decomposition~\cite{joung2021learning}. A render-and-compose calibration then transfers these predictions to the native frame of an unseen CTA. This allows the trained network to be applied with fixed weights, without patient-specific adaptation or explicit 3D preregistration. Our contributions are as follows:

\begin{enumerate}
    \item \textbf{Compositional frame transfer without preregistration.}
    We introduce a render-and-compose calibration that estimates the rigid transform between a patient's native CTA coordinate frame and the learned template frame from a single synthetic projection and one network forward pass. This enables a population-trained pose estimator to operate directly on native, unregistered CTA volumes without per-patient network training or explicit inter-volume 3D preregistration.

    \item \textbf{Population-trained, image-gated pose refinement.}
    We develop a residual refinement module shared across patients that predicts pose corrections from paired DSA MAP and patient-CTA renderings. An image-similarity gate retains only corrections that improve the observed alignment, providing a fast optimization-free operating point while limiting harmful recursive updates.

    \item \textbf{Rapid test-time registration.}
    We combine GeoPose initialization with low-budget NAdam optimization~\cite{dozat2016incorporating} of the GeoReg objective. The resulting framework preserves image-driven test-time finetuning while reducing the optimization budget from 400 to 25 iterations, enabling a registration timeframe of $\sim$2 seconds.

    \item \textbf{Experimental validation.}
    On a held-out test set of 80 DSA acquisitions, GeoPose achieves an mPCD of 5.8\,mm without test-time optimization while using fixed population-level weights, compared with 14.5\,mm for the best-performing baseline.
\end{enumerate}

\section{Method}
\label{sec:method}
GeoPose uses two population-trained networks with fixed test-time weights: \textit{GeoPose-Init} predicts a C-arm pose, and \textit{GeoPose-Refine} predicts a residual pose correction. Render-and-compose calibration maps the canonical-frame prediction to the native CTA frame, and similarity-gated refinement subsequently initializes a short GeoReg optimization. An overview of the method is presented in Fig.~\ref{fig:main}.

 \subsection{Direct CTA-to-DSA registration}
\label{sec:method formulation}
Let $V^\nat:\mathbb{R}^3\rightarrow\mathbb{R}$ denote a three-dimensional CTA volume represented in its native coordinate frame. For each view $v\in~\{\mathrm{LAT^-},\mathrm{PA},\mathrm{LAT}^+\}$, a DSA sequence consists of a set of temporal images $\{I_{\DSA}(t)\}_{t=1}^{T}$, where $T$ denotes the number of acquisition frames. As in GeoReg~\cite{van2026georeg}, a maximum-intensity projection over the temporal window is calculated to bridge the CTA-to-DSA domain gap:
\begin{equation}
\label{eq:dsa_map}
I_{\MAP}(\mathbf{p})
=
\mathcal{M}(\mathbf{p})
\max_{t\in\{1,\ldots,T\}}I_{\DSA}(\mathbf{p}, t),
\end{equation}
where $\mathbf{p}$ denotes a detector pixel and $\mathcal{M}$ is a coarse mask defining the valid cranium region. This temporal maximum-intensity projection recovers a silhouette of the subtracted \mbox{X-ray} sequence.
 
A C-arm pose is represented by a rigid camera-to-volume transformation $\mat{T}_v\in\mathrm{SE}(3)$. Given CTA volume $V^\nat$, pose $\mat{T}_v$, and camera intrinsic matrix $\mat{K}_v$, a DRR is generated by a differentiable rendering operator~\cite{gopalakrishnan2022fast}:
\begin{equation}
\label{eq:drr}
I_{\DRR,v}
=
\mathcal{R}
\left(
    V^\nat,
    \mat{T}_v;
    \mat{K}_v
\right),
\end{equation}
where $\mathcal{R}(\cdot)$ integrates the CTA intensities along rays cast from the virtual X-ray source toward the detector~\cite{siddon1985fast}. The objective is to estimate a pose $\mat{T}_v$ for which $I_{\DRR,v}$ aligns with the corresponding DSA MAP $I_{\MAP}$.

\begin{wrapfigure}[15]{l}{0.5\textwidth}
     \centering
     \includegraphics[width=0.5\textwidth, trim=1cm 0 0 0]{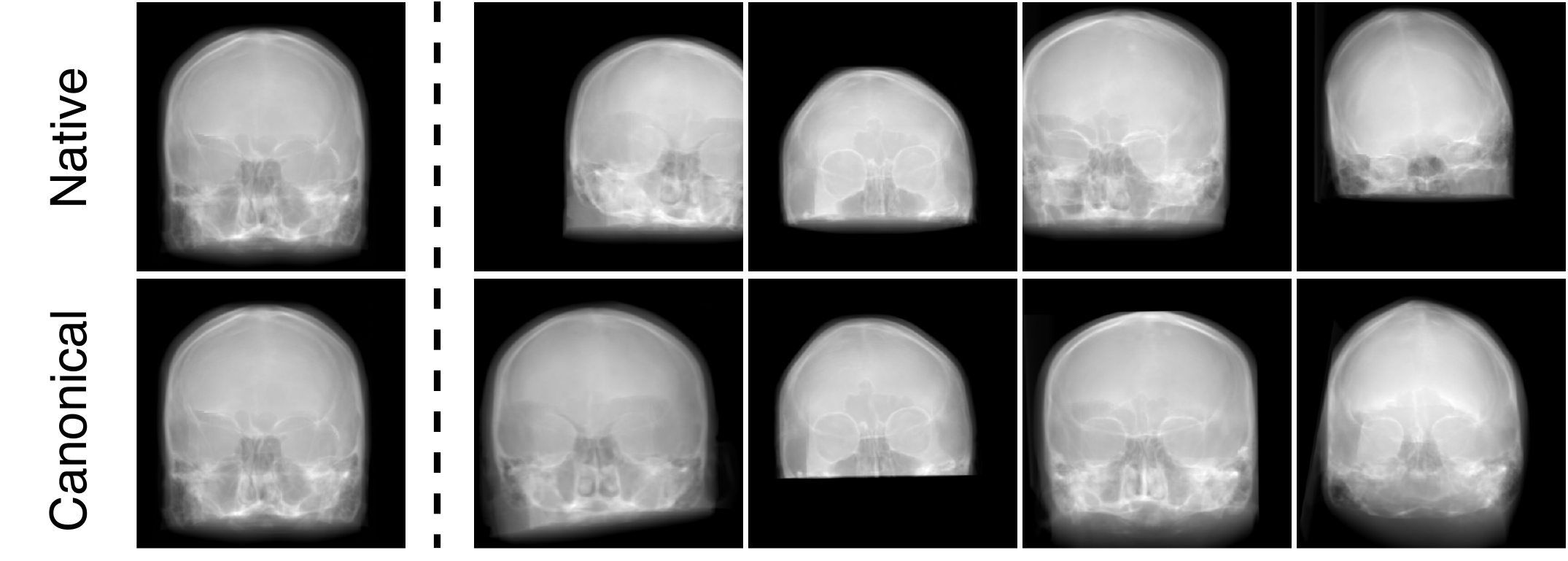}
     \caption{Native CTA (top) and canonical template-aligned (bottom) renders at the fixed posteroanterior isopose for four subjects from the ISLES'24 dataset. Aligning all training volumes to a shared frame (left) makes pose estimation a generalized function across subjects rather than per-patient regression.}
     \label{fig:canonical}
\end{wrapfigure}

\subsection{Population-trained pose estimation}
\label{sec:method geopose}

Training 3D pose estimation exclusively on a population implies \emph{pose--content decomposition}~\cite{joung2021learning}. Since every training volume is rendered in a shared template frame over a common pose distribution, the estimator absorbs patient-specific anatomy while reading out pose in a coordinate frame whose axes carry the same meaning for every volume. Pose estimation thus becomes a query within a canonical frame rather than a per-patient regression problem~(see Fig.~\ref{fig:canonical}), permitting the fixed-weight networks to transfer to an unseen CTA through the frame-transfer calibration described in Section~\ref{sec:method inference}.

\begin{figure*}
     \includegraphics[trim=0cm 1.5cm 1cm 1cm, clip, width=1.\textwidth]{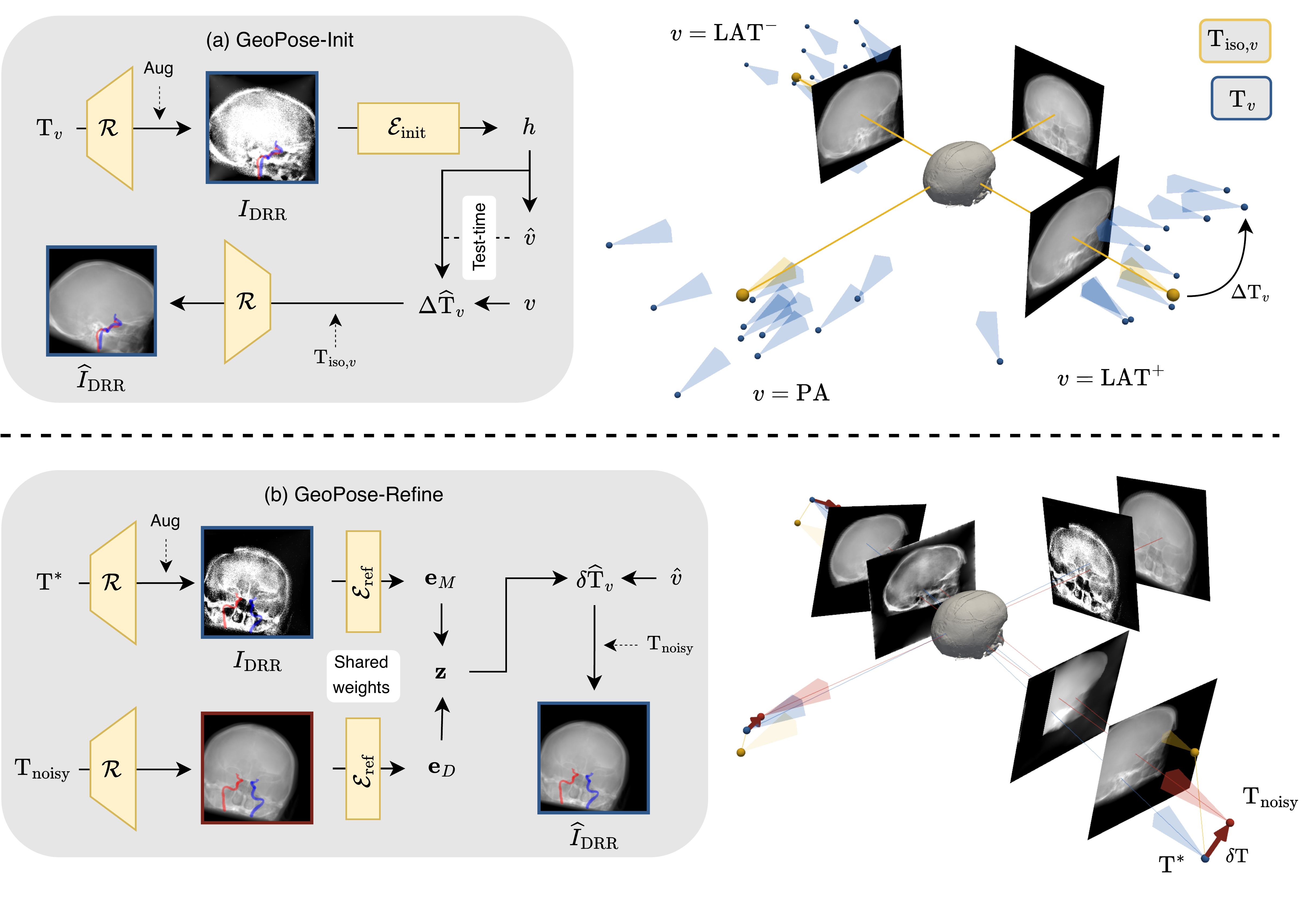}
     \caption{GeoPose training overview. \textit{(a) GeoPose-Init}: a training CTA is rendered at a sampled pose $\mathbf{T}_v$ and intensity-augmented to resemble a DSA MAP image, and subsequently passed through the encoder $\mathcal{E}_{\mathrm{init}}$. The network both classifies the originating view $v$  and regresses a residual $\Delta\widehat{\mathbf{T}}_v$ about the view-dependent isopose $\mathbf{T}_{\mathrm{iso},v}$. At test time, the predicted view class $\widehat{v}$ is used to derive the isopose. The right panel shows the isopose anchors $\mathbf{T}_{\mathrm{iso},v}$ (yellow) and the distribution of sampled training poses $\mathbf{T}_v$ (blue) for the three view classes ($v = \mathrm{LAT}^{-}, \mathrm{PA}, \mathrm{LAT}^{+}$) relative to the cranium. \textit{(b)~GeoPose-Refine}: a clean DRR rendered at a perturbed pose $\mathbf{T}_{\mathrm{noisy}}$ and an appearance-augmented, MAP-like DRR rendered at the optimal pose $\mathbf{T}^{*}$ are passed through a shared encoder $\mathcal{E}_{\mathrm{ref}}$ to obtain pooled embeddings $\mathbf{e}_M$ and $\mathbf{e}_D$. Their fused comparative embedding $\mathbf{z}$ is used to regress a residual transformation $\delta\hat{\mathbf{T}}_v$ that maps the perturbed pose back toward $\mathbf{T}^{*}$. The right panel illustrates the geometric relationship between $\mathbf{T}^{*}$, $\mathbf{T}_{\mathrm{noisy}}$, and the sampled perturbation $\delta\mathbf{T}$. Both models are trained using a combination of image similarity, vascular overlap, and geodesic losses.}
     \label{fig:training}
\end{figure*}

This decomposition is maintained through two population-trained networks: \emph{GeoPose-Init} and \emph{GeoPose-Refine}. GeoPose-Init directly regresses a six-degree-of-freedom C-arm pose from a MAP-like image, while GeoPose-Refine regresses a residual pose correction from a pair of MAP-like and rendered images, a cascaded refinement strategy inspired by LXPose~\cite{facente2026toward}. Both networks are optimized exclusively on synthetic renderings of a training population. Fig.~\ref{fig:training} provides an overview of the training procedure for both networks.

\subsubsection{GeoPose-Init}
\label{sec:method geopose init}

GeoPose-Init represents each C-arm pose as a residual relative to one of three view-dependent isoposes in a shared canonical coordinate frame. Let $v$ denote the view class defined in Section \ref{sec:method formulation}, where $\mathrm{LAT}^{-}$ and $\mathrm{LAT}^{+}$ distinguish the two opposing lateral C-arm orientations. Each class is assigned a fixed canonical rotation anchor:
\begin{equation}
\label{eq:view-anchors}
\overline{\mathbf{r}}_{v}
=
\begin{cases}
(-\frac{\pi}{2},0,0), & v=\mathrm{LAT}^{-},\\
(0,0,0),              & v=\mathrm{PA},\\
(+\frac{\pi}{2},0,0), & v=\mathrm{LAT}^{+}.
\end{cases}
\end{equation}
All three anchors share the isocentric translation $\overline{\mathbf{t}}=(0,650,0)^\top$ mm. Together, $\overline{\mathbf{r}}_v$ and $\overline{\mathbf{t}}$ define the view-dependent isopose $\mat{T}_{\mathrm{iso},v}$. GeoPose-Init therefore estimates a pose by predicting (i) the view class and (ii) a six-degree-of-freedom residual relative to the corresponding isopose.

Given an input projection $I$, an image-only classification head predicts logits for the three view classes and selects $\widehat{v}$. A pose-regression head separately predicts a rotational residual $\Delta\mathbf{r}$ and a translational residual $\Delta\mathbf{t}$. Both heads operate on the shared image representation
\begin{equation}
\mathbf{h}=f_{\theta}(I),
\end{equation}
produced by a one-channel ResNet encoder~\cite{he2016deep}. The pose-regression head is additionally conditioned on a learned embedding of the selected view class:
\begin{equation}
\label{eq:geopose-init-regression}
(\Delta\mathbf{r},\Delta\mathbf{t})
=
g_{\theta}\!\left(
    \left[\mathbf{h},\mathbf{e}_{\widetilde{v}}\right]
\right),
\end{equation}
where $\widetilde{v}=v$ during training and $\widetilde{v}=\widehat{v}$ at inference. Thus, the known view class of a synthetic training projection conditions the regression head, whereas an unseen input is processed using the class predicted from its image appearance. The classification head does not receive the view embedding.

The selected rotation anchor and the predicted residuals define the estimated pose:
\begin{equation}
\label{eq:geopose-init-pose}
\widehat{\mathbf{r}}
=
\overline{\mathbf{r}}_{\widetilde{v}}
+
\Delta\mathbf{r},
\qquad
\widehat{\mathbf{t}}
=
\overline{\mathbf{t}}
+
\Delta\mathbf{t}.
\end{equation}
The rotation is composed in Euler space, and $(\widehat{\mathbf{r}},\widehat{\mathbf{t}})$ is used to construct the predicted rigid pose transform $\widehat{\mat{T}}_v\in\mathrm{SE}(3)$. This view-anchored parameterization restricts the regression target to a residual about the nearby C-arm orientation rather than requiring the network to regress over the full pose space.

To train GeoPose-Init across patients, each training CTA is rigidly aligned to the shared canonical coordinate frame. For patient $i$, a pose $\mat{T}_v\in\mathrm{SE}(3)$ is sampled around one of the three view-dependent isoposes and used to render a synthetic projection:
\begin{equation}
\label{eq:geopose training render}
I_\DRR
=
\mathcal{R}
\left(
    V_i^\can,
    \mat{T}_v;
    \mat{K}_0
\right),
\end{equation}
where $V_i^\can$ denotes the canonical-frame CTA and $\mat{K}_0$ denotes the fixed projection geometry used during training. Because the pose and its associated isopose are sampled directly, the view class $v$ is known for every synthetic projection.

Given the resulting pose prediction $\widehat{\mat{T}}_v$, a second DRR is rendered at the predicted pose:
\begin{equation}
\widehat{I}_\DRR
=
\mathcal{R}
\left(
    V_i^\can,
    \widehat{\mat{T}}_v;
    \mat{K}_0
\right).
\end{equation}
The network is then trained by minimizing
\begin{equation}
\label{eq:geopose loss}
\begin{split}
\mathcal{L}_{\mathrm{init}}
={}&
-\operatorname{mNCC}(I_\DRR,\widehat{I}_\DRR)
+
\lambda_{\mathrm{geo}}
d_{\mathrm{geo}}
\left(
    \mat{T}_v,\widehat{\mat{T}}_v
\right)
\\&+
\lambda_{\mathrm{art}}
\mathcal{L}_{\mathrm{Dice+mPD}}
\left(
    M,\widehat{M}
\right)
+
\lambda_{\mathrm{view}}
\mathcal{L}_{\mathrm{CE}}
\left(
    v,\widehat{v}
\right).
\end{split}
\end{equation}
Here, $\operatorname{mNCC}$ denotes multiscale normalized cross-correlation, and $d_{\mathrm{geo}}$ combines rotation and translation error in a common physical coordinate system~\cite{gopalakrishnan2024intraoperative}. The masks $M$ and $\widehat{M}$ denote the projected vascular occupancy at the sampled and predicted poses, respectively. $\mathcal{L}_{\mathrm{Dice+mPD}}$ combines a Dice term~\cite{sudre2017generalised} with a projection-distance penalty~\cite{facente2026toward} to promote vascular overlap, while $\mathcal{L}_{\mathrm{CE}}$ is the cross-entropy loss used to train the view classifier. After training on synthetic projections from the canonicalized population, GeoPose-Init is applied with fixed weights to both CTA-derived DRRs and real DSA maximum-intensity projections.

\subsubsection{GeoPose-Refine}
\label{sec:method georefine}
To further improve a pose estimate without invoking gradient-based optimization, a second population-trained network is introduced that predicts a residual rigid transformation from the discrepancy between a MAP-like image and a rendering of the current pose estimate. A secondary ResNet encoder, shared between the two inputs, processes both images individually, resulting in pooled feature representations $\mathbf{e}_M$ and $\mathbf{e}_D$. A comparative embedding is formed as
\begin{equation}
\label{eq:refine features}
\mathbf{z}
=
\left[
    \mathbf{e}_M,\,
    \mathbf{e}_D,\,
    \mathbf{e}_M-\mathbf{e}_D,\,
    \left|\mathbf{e}_M-\mathbf{e}_D\right|,\,
    \mathbf{e}_M\odot\mathbf{e}_D
\right],
\end{equation}
where $\odot$ denotes element-wise multiplication. After linear fusion, $\mathbf{z}$ is concatenated with a learned embedding of the view~$\widehat{v}$. From this, the network predicts an axis-angle rotation $\delta\mathbf{\widehat{r}}$ and translation $\delta\mathbf{\widehat{t}}$, which together define a residual transformation~$\mat{\delta}\widehat{\mat{T}}\in\mathrm{SE}(3)$.

The refiner is trained using synthetic pairs where, for an optimal pose $\mat{T}^{*}$, a perturbation $\mat{\delta}\mat{T}$ is sampled which constructs
\begin{equation}
\label{eq:refine perturbation}
\mat{T}_{\mathrm{noisy}}
=
\mat{T}^{*}\mat{\delta}\mat{T}.
\end{equation}
For each synthetic pair, the first network input is a DRR rendered at $\mat{T}^{*}$ and augmented with appearance corruptions to emulate a MAP-like image, whereas the second input is a clean DRR rendered at $\mat{T}_{\mathrm{noisy}}$. Thus, both inputs are synthetically generated during training, while retaining the appearance discrepancy that the refiner must resolve at inference. Since $\mat{\delta}\mat{T}$ maps the optimal pose to the perturbed pose, the corrected prediction is obtained by composing the inverse residual:
\begin{equation}
\label{eq:refine correction}
\widehat{\mat{T}}^{*}
=
\mat{T}_{\mathrm{noisy}}
{\mat{\delta}}\widehat{\mat{T}}^{-1}.
\end{equation}
The refinement module training objective follows Eq.~\ref{eq:geopose loss}, omitting the view classification term:
\begin{equation}
\label{eq:refine loss}
\begin{split}
\mathcal{L}_{\mathrm{ref}}
={}&
-\operatorname{mNCC}
\left(
    I_\DRR^*,
    \widehat{I}_\DRR^*
\right)
+
\lambda_{\mathrm{geo}}
d_{\mathrm{geo}}
\left(
    \mat{T}^{*},
    \widehat{\mat{T}}^{*}
\right)
\\&
+
\lambda_{\mathrm{art}}
\mathcal{L}_{\mathrm{Dice+mPD}}
\left(
    M^*,
    \widehat{M}^*
\right).
\end{split}
\end{equation}
The shared encoder is initialized from the trained GeoPose-Init model and subsequently fine-tuned jointly with the newly initialized fusion, view-embedding, and residual-prediction layers on synthetic pose perturbations. At inference, all parameters remain fixed and the refiner is applied without gradient-based optimization or patient-specific adaptation.

\subsection{Test-time inference}
\label{sec:method inference}
GeoPose-Init and GeoPose-Refine are trained across a training population in a predetermined canonical framework. As such, applying them to an unregistered patient CTA requires transferring predictions from the learned canonical frame into the patient's native CTA frame, deciding when a proposed refinement should be applied, and, optionally, adapting the resulting pose to the observed DSA pair through direct image similarity-driven optimization.

\subsubsection{Projection-space calibration}
GeoPose is trained to predict poses relative to a shared canonical coordinate frame. At inference, however, the estimated C-arm pose must be expressed relative to the native coordinate frame of the patient CTA. Explicitly registering each new CTA to the training template would resolve this discrepancy, but would introduce a separate 3D registration problem into the time-critical workflow. Instead, we estimate the required frame transfer from a single synthetic projection of the native CTA.

Let $\mat{A}\in\mathrm{SE}(3)$ denote the transformation from canonical to native volume coordinates. Since DiffDRR represents poses as camera-to-volume transformations \cite{gopalakrishnan2022fast}, the equivalent pose in the canonical volume frame is obtained by left-composition with $\mat{A}^{-1}$. Consequently, the same projection satisfies
\begin{equation}
\mathcal{R}
\left(
    V^{\mathrm{nat}},
    \mat{T};
    \mat{K}
\right)
=
\mathcal{R}
\left(
    V^{\mathrm{can}},
    \mat{A}^{-1}\mat{T};
    \mat{K}
\right).
\label{eq:frame_rendering_identity}
\end{equation}

We exploit this identity by rendering the native CTA at a known isocentric PA calibration pose $\mat{T}_{\mathrm{cal}}^{\mathrm{nat}}$:
\begin{equation}
I_{\mathrm{cal}}
=
\mathcal{R}
\left(
    V^{\mathrm{nat}},
    \mat{T}_{\mathrm{cal}}^{\mathrm{nat}};
    \mat{K}_0
\right).
\label{eq:calibration_render}
\end{equation}
Although $I_{\mathrm{cal}}$ is rendered at $\mat{T}_{\mathrm{cal}}^{\mathrm{nat}}$ in the native frame, the same projection corresponds to $\mat{A}^{-1}\mat{T}_{\mathrm{cal}}^{\mathrm{nat}}$ in the canonical frame. Processing it with GeoPose-Init therefore gives
\begin{equation}
\widehat{\mat{T}}_{\mathrm{cal}}^{\mathrm{can}}
=
g_{\theta}
\left(
    \left[
        f_{\theta}(I_{\mathrm{cal}}),
        \mathbf{e}_{\mathrm{PA}}
    \right]
\right)
\approx
\mat{A}^{-1}\mat{T}_{\mathrm{cal}}^{\mathrm{nat}}.
\label{eq:calibration_prediction}
\end{equation}
The canonical-to-native frame transfer can then be recovered by composition:
\begin{equation}
\widehat{\mat{A}}
=
\mat{T}_{\mathrm{cal}}^{\mathrm{nat}}
\left(
    \widehat{\mat{T}}_{\mathrm{cal}}^{\mathrm{can}}
\right)^{-1}.
\label{eq:frame_transfer}
\end{equation}

For a DSA MAP image $I_{\mathrm{MAP},v}$, let $\widehat{\mat{T}}_{v}^{\mathrm{can}}$ denote the pose predicted by GeoPose-Init in the canonical frame. Its corresponding pose in the native CTA frame is
\begin{equation}
\widehat{\mat{T}}_{v,\mathrm{init}}^{\mathrm{nat}}
=
\widehat{\mat{A}}\,
\widehat{\mat{T}}_{v}^{\mathrm{can}}
=
\mat{T}_{\mathrm{cal}}^{\mathrm{nat}}
\left(
    \widehat{\mat{T}}_{\mathrm{cal}}^{\mathrm{can}}
\right)^{-1}
\widehat{\mat{T}}_{v}^{\mathrm{can}}.
\label{eq:native_pose_composition}
\end{equation}
The calibration requires only one additional DRR and one GeoPose-Init evaluation per patient. The resulting frame transfer $\widehat{\mat{A}}$ is shared by all PA and LAT predictions for that patient.

\subsubsection{Gated residual refinement}
\label{sec:method refinement}
\label{sec:method greedy refinement}
The calibrated pose in Eq.~\ref{eq:native_pose_composition} provides an initialization in the patient-native CTA frame. GeoPose-Refine (Section~\ref{sec:method georefine}) may be applied to this initialization by substituting the real $I_{\MAP,v}$ and a rendering of the current pose estimate for the MAP-like and render-like inputs used during training. For a pose $\widehat{\mat{T}}_{v,k}^{\nat}$ at refinement step $k$, we render
\begin{equation}
\label{eq:refine render}
I_{\DRR,v,k}
=
\mathcal{R}
\left(
    V^\nat,
    \widehat{\mat{T}}_{v,k}^{\nat};
    \mat{K}_0
\right)
\end{equation}
from the patient's native CTA and evaluate the trained refiner on $(I_{\MAP,v}, I_{\DRR,v,k})$ to obtain a residual transformation $\widehat{\mat{\delta}}_{v,k}$.

To facilitate iterative pose refinement, GeoPose-Refine is applied under a greedy acceptance criterion based on image similarity~(mNCC). Given the current pose $\widehat{\mat{T}}_{v,k}^{\nat}$, the refiner proposes
\begin{equation}
\label{eq:refine candidate}
\widehat{\mat{T}}_{v,k+1}^{\nat}
=
\widehat{\mat{T}}_{v,k}^{\nat}
\widehat{\mat{\delta}}_{v,k}^{-1}.
\end{equation}
The candidate pose is retained only while the image similarity between the newly rendered $I_{\DRR,v,k+1}$ at the updated pose and $I_{\MAP,v}$ increases, and is permitted at most $N_{\text{ref}}$ refinement steps.

\subsubsection{Test-time optimization}
\label{sec:method optimization}
Although GeoPose-Init and GeoPose-Refine provide rapid pose estimates, pose convergence is not guaranteed. To adapt the registration directly to the observed DSA pair, the resulting pose estimates are used to initialize a short differentiable optimization using GeoReg.
 
For each view, the patient CTA is rendered using the true acquisition geometry $\mat{K}_v$. The image-similarity loss is defined as
\begin{equation}
\label{eq:tto ncc}
\mathcal{L}_{\mathrm{NCC}}^{v}
=
-
\operatorname{mNCC}
\left(
    I_{\mathrm{MAP},v},
    \mathcal{R}
    \left(
        V^\nat,
        \mat{T}_v;
        \mat{K}_v
    \right)
\right).
\end{equation}
Additionally, a lightweight silhouette term compares the DSA cranium mask $\mathcal{M}_v$ with the projected CTA skull channel $\widehat{\mathcal{M}}_v(\mat{T}_v)$:
\begin{equation}
\label{eq:tto dice}
\mathcal{L}_{\mathrm{mask}}^{v}
=
\mathcal{L}_{\mathrm{GDL}}
\left(
    \mathcal{M}_v,
    \sigma
    \left(
        \widehat{\mathcal{M}}_v(\mat{T}_v)
    \right)
\right),
\end{equation}
where $\mathcal{L}_{\mathrm{GDL}}$ is the generalized Dice loss~\cite{sudre2017generalised} and $\sigma$ is the sigmoid activation function. Both view poses are optimized jointly, resulting in the complete test-time objective:
\begin{equation}
\label{eq:tto objective}
\mathcal{L}_{\mathrm{TTO}}
=
\sum_{v\in\{\mathrm{PA},\mathrm{LAT}^{-/+}\}}
\left[
    \alpha
    \mathcal{L}_{\mathrm{NCC}}^{v}
    +
    (1-\alpha)
    \mathcal{L}_{\mathrm{mask}}^{v}
\right].
\end{equation}
As $\mathbf{T}_{v,\text{ref}}^{\nat}$ already lies near the true pose, the optimization refines the existing estimate through a short NAdam run of 25 iterations without additional geodesic constraints. This setting is used to produce the final per-view poses $\mathbf{T}_{v,f}$.

\section{Experiments}
\label{sec:experiments}

The proposed method is evaluated on CTA-to-DSA registration for acute ischemic stroke, where performance is assessed both with and without test-time optimization. Section~\ref{sec:data} describes the datasets, while Section~\ref{sec:implementation} provides methodological implementation details. Baselines and ablations are described in Section~\ref{sec:baselines}, and the evaluation protocol is defined in Section~\ref{sec:evaluation}.

\begin{table*}[t]
\centering
\caption{Instant DSA pose-estimation performance on the reserved ISLES'24 test set ($N_{\text{DSA}}=80$), without test-time optimization. Normalized cross-correlation (NCC), carotid mean projected centerline distance (mPCD), and centerline Dice (clDice) are reported. Runtime covers the complete initialization stage for one biplanar pair, measured on an NVIDIA H100 GPU. Results are presented as mean (standard deviation). Avg. denotes the patient-level average across PA and LAT views, while a $\dagger$ indicates statistical significance ($p\leq0.0015$) compared to our greedy refinement model using the Wilcoxon signed-rank test. --T denotes training with TopCoW data, while an asterisk ($^*$) denotes evaluation on the broader train+test cohort.}
\label{tab:instant_pose}
\resizebox{\textwidth}{!}{
\begin{tabular}{lcccccccccc}
\toprule
& \multicolumn{3}{c}{NCC $\uparrow$}
& \multicolumn{3}{c}{mPCD $\downarrow$ (mm)}
& \multicolumn{3}{c}{clDice $\uparrow$}
& Runtime $\downarrow$ (ms) \\
\cmidrule(lr){2-4}
\cmidrule(lr){5-7}
\cmidrule(lr){8-10}
& PA & LAT & Avg. & PA & LAT & Avg. & PA & LAT & Avg. \\
\midrule
Native init. & 0.57 (0.16) & 0.34 (0.20) & 0.46 (0.13) & 14.9 (7.0) & 28.4 (18.0) & 21.6 (9.6) & 0.14 (0.11) & 0.09 (0.11) & 0.11 (0.08) & - \\

xvr$^\dagger$ & 0.71 (0.11) & 0.66 (0.09) & 0.68 (0.08) & 19.7 (10.0) & 21.4 (13.7) & 20.6 (9.2) & 0.12 (0.09) & 0.15 (0.13) & 0.13 (0.09) & 27.4 (2.3) \\

LXPose (net1)$^\dagger$ & 0.75 (0.16) & 0.70 (0.13) & 0.73 (0.13) & 8.7 (5.8) & 25.9 (21.4) & 17.3 (12.8) & 0.27 (0.17) & 0.12 (0.18) & 0.19 (0.13) & \textbf{20.9} (2.3) \\

LXPose (net1 + net2)$^\dagger$ & 0.77 (0.16) & 0.75 (0.13) & 0.76 (0.13) & 7.0 (6.4) & 21.9 (20.8) & 14.5 (12.4) & 0.38 (0.18) & 0.18 (0.20) & 0.28 (0.15) & 49.8 (6.7) \\

\midrule

GeoPose-Init & 0.78 (0.13) & 0.72 (0.14) & 0.75 (0.12) & 9.4 (10.2) & 22.0 (15.5) & 15.7 (10.7) & 0.34 (0.22) & 0.13 (0.18) & 0.24 (0.16) & 21.8 (2.3) \\

GeoPose-Init + Refine ($\times1$) & 0.81 (0.09) & 0.81 (0.11) & 0.81 (0.09) & 5.9 (4.9) & 7.8 (4.2) & 6.9 (4.3) & 0.46 (0.22) & 0.32 (0.13) & 0.39 (0.15) & 52.1 (6.9) \\

GeoPose-Init + Refine (greedy) & \textbf{0.83} (0.08) & \textbf{0.84} (0.09) & \textbf{0.84} (0.07) & \textbf{4.6} (2.4) & \textbf{7.0} (5.0) & \textbf{5.8} (2.8) & \textbf{0.51} (0.21) & \textbf{0.40} (0.13) & \textbf{0.45} (0.15) & 147.0 (39.0) \\

\midrule \midrule
\textit{Out-of distribution training} \\
GeoPose-Init + Refine (greedy) --T & 0.81 (0.10) & 0.81 (0.07) & 0.81 (0.07) & 7.7 (3.8) & 9.2 (4.3) & 8.5 (3.3) & 0.30 (0.15) & 0.27 (0.13) & 0.28 (0.10) & 122.1 (42.2) \\
GeoPose-Init + Refine (greedy) --T$^*$ & 0.76 (0.13) & 0.75 (0.14) & 0.76 (0.13) & 10.9 (8.7) & 12.6 (11.0) & 11.7 (7.6) & 0.26 (0.16) & 0.25 (0.16) & 0.25 (0.13) & 122.1 (42.2) \\

\bottomrule
\end{tabular}
}
\end{table*}

\subsection{Data}
\label{sec:data}
Two cerebrovascular stroke datasets were utilized in this study: the ISLES'24 challenge dataset~\cite{de2024isles} and the TopCoW dataset~\cite{yang2025benchmarking}. All primary experiments were conducted on the ISLES'24 dataset, which was separated into a 70/10/20 split for training, validation, and testing. To assess out-of-distribution generalization, TopCoW was additionally utilized as an alternative training source, allowing the combined ISLES'24 train and test set to serve as a hold-out evaluation set.

For the \textbf{ISLES'24} dataset~\cite{de2024isles}, we utilize the publicly available CTA volumes acquired as part of standard stroke assessment protocols. A subset of 99 CTA volumes was matched with an internal collection of biplanar DSA acquisitions obtained during endovascular procedures at the TUM University Hospital, for which DSA sequences consisted of pairs of PA and $\text{LAT}^{-/+}$ projections acquired using standard neurovascular acquisition protocols. The available subset has a median age of 79 (range 37--98) and includes 49 male patients. CTA scans were reconstructed to a median in-plane isotropic voxel size of 0.60 mm (range 0.31--0.97 mm) with a median slice thickness and increment of 0.40 mm (range 0.39--1.00~mm). 

Each patient contributed a median of 4 DSA series (range 2--4; 369 total), acquired as biplanar PA/LAT pre- and post-intervention pairs with primary angulations clustered near $0^\text{o}$ and $-90^\text{o}$ (median $-14.7^\text{o}$, range $-115.5^\text{o}$ to $84.0^\text{o}$). Series were acquired at a median source-to-detector distance of 1086 mm (range 970--1302 mm), and imaging series comprises a median of 14 frames (range 3--26) at a detector matrix between $870\times870$ and $1920\times1904$ pixels (median $1428\times1428$).

The \textbf{TopCoW} dataset~\cite{yang2025benchmarking} includes a cohort of 125 patients with stroke-related neurological disorders, for which CTA images and corresponding Circle of Willis (CoW) annotations are made publicly available. Images have a median in-plane voxel size of 0.43 mm (range 0.34--0.63 mm) and a mean slice thickness of 0.72 mm. Note that TopCoW training utilizes the CoW for the $\mathcal{L}_{\mathrm{Dice+mPD}}$ loss component.

\subsection{Implementation details}
\label{sec:implementation}
\subsubsection{Training-time canonicalization} For canonical-frame training, all images were rigidly registered to a single template, arbitrarily defined by the first patient in the ISLES'24 dataset (\texttt{sub-stroke0001}). Using FireANTs~\cite{jena2024fireants}, a multi-scale affine transform was optimized between the HD-BET brain masks~\cite{isensee2019automated} of the moving and fixed image. Its linear part was then projected onto $\mathrm{SO}(3)$ via SVD polar decomposition, thus discarding scale and shear. The resulting rigid transform was applied to the moving CTA image and any optional label maps.

\subsubsection{GeoPose} Synthetic DRRs were rendered on a $256\times256$ detector with $1.2\,\mathrm{mm}$ pixel spacing and a source-to-detector distance of $1020\,\mathrm{mm}$. Poses were sampled equally between PA and LAT views, with Gaussian rotational and translational deviations of $0.2\,\mathrm{rad}$ and $(25,75,25)\,\mathrm{mm}$ respectively, about the view-dependent isoposes. GeoPose-Refine perturbations were calibrated to the measured per-axis residual errors of GeoPose-Init.

Training projections were heavily augmented as proposed by Facente~\textit{et al.}~\cite{facente2026toward}, specifically including plasma~\cite{nicolaou2022tormentor} and bilateral transformations~\cite{van2026georeg}. Both networks employ ResNet-34 backbones~\cite{he2016deep} with 16-dimensional view embeddings. Networks were trained for 400 epochs using AdamW~\cite{loshchilov2017decoupled} with weight decay \(10^{-5}\), cosine learning-rate decay~\cite{loshchilov2016sgdr}, and an initial learning rate of \(1.5\times10^{-4}\). The batch size was four, with gradient accumulation over eight batches for GeoPose-Init and four for GeoPose-Refine. Optimization weights are set to $\lambda_{\mathrm{geo}}=0.01$ and $\lambda_{\mathrm{art}}=0.1$, with GeoPose-Init additionally using a view-classification weight of \mbox{$\lambda_{\mathrm{view}}=1$}.

\subsubsection{Test-time inference and optimization}
Greedy refinement was limited to $N_{\text{ref}}=5$ steps and terminated at the first update that did not improve mNCC. Test-time optimization used NAdam~\cite{dozat2016incorporating} for 25 iterations with \(\alpha=0.5\), a learning rate of \(10^{-4}\), and a OneCycle schedule peaking at \(10^{-2}\). We refer to GeoReg~\cite{van2026georeg} for additional implementation details.

\subsection{Baseline methods and ablations}
\label{sec:baselines}
We compare two classes of methods. Optimization-only baselines use either direct DiffDRR-based registration~\cite{gopalakrishnan2022fast} or GeoReg~\cite{van2026georeg}, both initialized from the native acquisition pose. For population-trained baselines, we compare against xvr~\cite{gopalakrishnan2025rapid} and LXPose~\cite{facente2026toward}, trained once on the same canonical population and evaluated with fixed weights using the projection-space calibration proposed in this work.

\begin{figure*}[t]
     \centering
     \includegraphics[width=\textwidth, trim=0 0.65cm 0 0, clip]{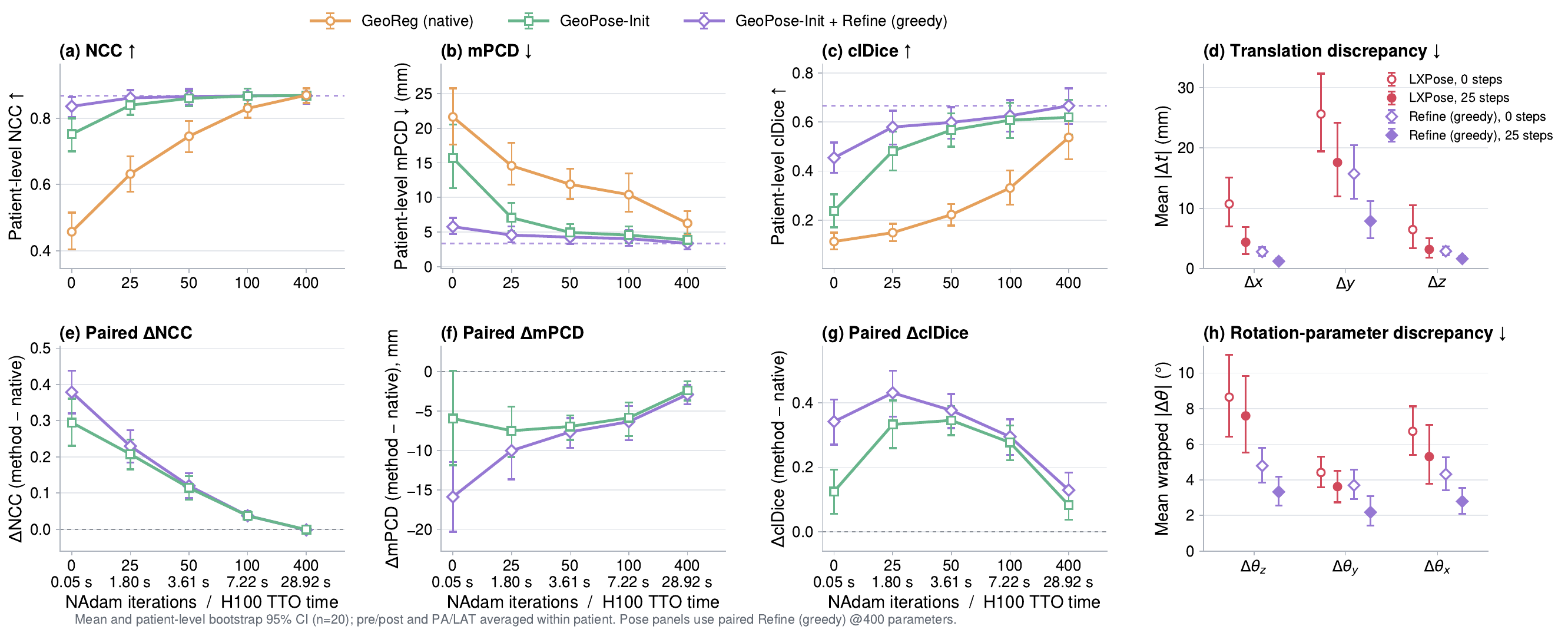}
     \caption{Test-time optimization (TTO) convergence and pose-parameter discrepancies on the ISLES'24 test set. \textit{Top}: (a) patient-level normalized cross-correlation (NCC), (b) carotid mean projected centerline distance (mPCD), and (c) centerline Dice (clDice) versus NAdam iteration count, and (d) absolute translation-component discrepancies relative to the long-TTO reference pose. \textit{Bottom}: (e--g) paired metric differences relative to the GeoReg native pose at matched iteration counts, and (h) absolute wrapped Euler-ZYX component discrepancies relative to the long-TTO reference. Pose panels compare LXPose (net1 + net2) and GeoPose-Init + Refine (greedy), before TTO and after independent 25-step schedules, against the paired GeoPose-Init + Refine (greedy) 400-step pose. Error bars show patient-level bootstrap 95\% confidence intervals ($N=20$), while pre/post and PA/LAT observations are averaged within patient.}
     \label{fig:tto}
\end{figure*}

GeoPose is evaluated after initialization alone, after one refinement step, and after greedy refinement with up to $N_{\text{ref}}=5$ steps. Each initialization is also evaluated as a starting point for GeoReg test-time optimization. To assess out-of-distribution performance, the complete pipeline is additionally trained on TopCoW and evaluated on the ISLES cohort without further adaptation.

Starting from a base GeoPose-Init model trained with geodesic and Dice supervision, we incrementally add the auxiliary view classifier, the view embedding supplied to the pose head, the view-dependent rotation-anchor loss, and the distance-based segmentation loss \(\mathcal{L}_{\text{mPD}}\). The final Dice+mPD GeoPose-Init model is subsequently fixed and used to compare Dice-only versus Dice+mPD refinement using either one update or the greedy refinement schedule. These comparisons isolate the contributions of image-based laterality prediction, view conditioning, view anchoring, mPD supervision, and iterative refinement.

\subsection{Evaluation}
\label{sec:evaluation}
Registration quality is assessed using NCC~\cite{lewis1995fast}, carotid mean projected centerline distance (mPCD, mm)~\cite{van2026georeg}, and carotid centerline Dice (clDice)~\cite{shit2021cldice,berger2025pitfalls}. mPCD measures the distance from the projected CTA carotid centerline to the corresponding segmented DSA centerline. Reference segmentations for carotid arteries were obtained by training two \mbox{nnU-Net} models~\cite{isensee2021nnu} on publicly available CTA~\cite{tiefenthaler2025shape} and DSA~\cite{zhang2025dsca} data.

For methods with an image-based view classifier, we report view classification accuracy over the \(\mathrm{LAT}^{-}\), PA, and \(\mathrm{LAT}^{+}\) classes. Additionally, pose errors in $\mathrm{SE}(3)$ are quantified w.r.t. the optimal pose identified by the best-performing model given 400 iterations of test-time optimization as a ground-truth surrogate. Metrics are reported separately for PA and LAT views and jointly across both views, where view measurements are first aggregated within each patient before cohort-level statistics are computed.

All learned initialization methods are first evaluated without test-time optimization to isolate pose-estimation performance. We then apply the same NAdam optimization for $\{25,50,100,400\}$ iterations to compare convergence speed from each initialization. Network inference and optimization runtimes are measured on an NVIDIA H100 GPU over ten patients after five unmeasured warm-up iterations, with CUDA synchronization at each timing boundary.

\section{Results}
\label{sec:res}

\subsection{Instant pose estimation}

Table~\ref{tab:instant_pose} presents the pose-estimation performance without test-time optimization. In general, the learned methods improve upon the native initialization, with LXPose performing better than xvr across the geometric registration metrics. GeoPose-Init performs similarly to the complete two-network LXPose configuration, while the addition of GeoPose-Refine results in a clear improvement across all metrics. Greedy refinement achieves the best overall performance, followed by refinement with a single update. Notably, refinement provides the largest improvements for LAT acquisitions, for which all initialization methods show lower geometric accuracy than for PA acquisitions. The complete greedy-refinement procedure requires approximately 0.15\,s per biplanar pair. Training GeoPose on TopCoW results in slightly lower performance on ISLES'24, with a further reduction when evaluating the broader ISLES'24 train and test set.

\subsection{Component ablation}

\begin{wraptable}[15]{r}{0.6\textwidth}
\vspace{-0.7cm}
\centering
\caption{
Cumulative GeoPose component ablation on the hold-out test set
($N_{\text{DSA}}=80$). Values are patient-level mean (standard deviation).
}
\label{tab:geopose_ablation}
\footnotesize
\setlength{\tabcolsep}{2.5pt}
\resizebox{0.6\columnwidth}{!}{
\begin{tabular}{@{}lcccc@{}}
\toprule
Configuration
& View acc. $\uparrow$
& NCC $\uparrow$
& mPCD $\downarrow$ (mm)
& clDice $\uparrow$ \\
\midrule

\multicolumn{5}{@{}l}{\emph{GeoPose-Init}} \\

Base
& N/A
& 0.74 (0.11)
& \textbf{12.6} (6.6)
& 0.22 (0.15) \\

+ View classification
& 98.8 (5.6)
& 0.73 (0.13)
& 14.1 (10.3)
& 0.23 (0.16) \\

+ View embedding
& \textbf{100.0} (0.0)
& 0.76 (0.11)
& 13.3 (7.8)
& 0.23 (0.13) \\

+ View-anchor loss
& 98.8 (5.6)
& \textbf{0.76} (0.09)
& 13.0 (7.2)
& \textbf{0.25} (0.16) \\

+ mPD loss
& \textbf{100.0} (0.0)
& 0.75 (0.12)
& 15.7 (10.7)
& 0.24 (0.16) \\

\midrule
\multicolumn{5}{@{}l}{\emph{GeoPose-Refine}} \\

+ Refine Dice ($\times1$)
& -- & 0.79 (0.11) & 8.6 (5.8) & 0.34 (0.16) \\

+ Refine Dice+mPD ($\times1$)
& -- & 0.81 (0.09) & 6.9 (4.3) & 0.39 (0.15) \\

+ Refine Dice (greedy)
& -- & 0.81 (0.10) & 8.0 (5.1) & 0.38 (0.16) \\

+ Refine Dice+mPD (greedy)
& --
& \textbf{0.84} (0.07)
& \textbf{5.8} (2.8)
& \textbf{0.45} (0.15) \\

\bottomrule
\end{tabular}
}
\vspace{2pt}

\parbox{0.6\columnwidth}{
\scriptsize
Rows are cumulative within each section. Refine rows use the final
Dice+mPD GeoPose-Init model. Refinement does not alter view classification, as all Refine rows inherit the final GeoPose-Init classifier.
}
\end{wraptable}

The contribution of the individual GeoPose components is presented in Table~\ref{tab:geopose_ablation}. Adding the view classifier, view embedding, and view-anchor loss modestly improves the corresponding GeoPose-Init configurations, and adding mPD supervision does not further improve GeoPose-Init. In contrast, mPD supervision consistently improves GeoPose-Refine compared with Dice supervision alone. This improvement is observed for both a single refinement update and greedy refinement. The combination of Dice and mPD supervision with greedy refinement achieves the best overall performance in the ablation.

\subsection{Test-time optimization}

Table~\ref{tab:tto_25} presents the registration performance after 25 iterations of test-time optimization. DiffDRR and GeoReg perform similarly when initialized from the native acquisition pose and are consistently outperformed by the learned initialization methods. Among these methods, xvr primarily improves NCC, whereas LXPose additionally improves mPCD and clDice. GeoPose-Init performs better than both LXPose configurations on geometric metrics, with further improvements obtained by GeoPose-Refine. After optimization, a single refinement update and greedy refinement perform similarly, although greedy refinement achieves the highest clDice.

The corresponding optimization curves are shown in Fig.~\ref{fig:tto}. GeoPose starts with better NCC, mPCD, and clDice and retains this advantage during the early optimization iterations. A similar trend is observed in the paired comparisons with native-frame GeoReg at matched iteration counts. After 400 iterations, both approaches reach the same average NCC, while GeoPose maintains better vascular alignment, with an mPCD of 3.4 versus 6.3\,mm and a clDice of 0.67 versus 0.54. The component-wise results further show that test-time optimization reduces the pose discrepancies for both learned initializers. The largest remaining discrepancies are observed for translation along the source--detector direction and the out-of-plane rotational components. Fig.~\ref{fig:qualitative_registration} shows the registration progression together with a direct application of the recovered pose, where an anatomically labeled CTA segmentation is forward projected onto the DSA MAP.

\section{Discussion}
\label{sec:discussion}

\begin{wraptable}[23]{l}{0.6\textwidth}
\vspace{-0.5cm}
\centering
\caption{Patient-level registration performance after 25 iterations of test-time optimization (TTO), initialized from each method's learned pose estimate. For each patient, results indicate mean (standard deviation) averaged across PA and LAT views. Normalized cross-correlation (NCC) and carotid mean projected centerline distance (mPCD) and clDice are reported. A $\dagger$ indicates statistical significance w.r.t. our greedy refinement method at 25 TTO steps ($p\leq0.01$). Bottom rows report a 400-iteration upper-bound reference for native-frame initialized GeoReg and for GeoPose (GP).}

\label{tab:tto_25}
\footnotesize
\setlength{\tabcolsep}{4pt}
\resizebox{0.6\columnwidth}{!}{
\begin{tabular}{lrccc}
\toprule
Initialization & Iter. & NCC $\uparrow$ & mPCD $\downarrow$ (mm) & clDice $\uparrow$ \\
\midrule
DiffDRR (native)$^\dagger$ & 25 & 0.62 (0.12) & 14.7 (6.7) & 0.15 (0.10) \\
GeoReg (native)$^\dagger$  & 25 & 0.63 (0.13) & 14.6 (7.2) & 0.15 (0.08) \\
xvr$^\dagger$        & 25 & 0.82 (0.05) & 13.3 (4.5) & 0.22 (0.13) \\
LXPose (net1)$^\dagger$      & 25 & 0.83 (0.07) & 8.8 (6.2) & 0.41 (0.18) \\
LXPose (net1 + net2)$^\dagger$ & 25 & 0.84 (0.07) & 8.0 (5.2) & 0.42 (0.18) \\
\midrule
GP-Init                 & 25 & 0.84 (0.07) & 7.1 (4.8) & 0.48 (0.18) \\
GP-Init + Refine ($\times1$) & 25 & \textbf{0.86} (0.06) & \textbf{4.5} (2.0) & 0.56 (0.15) \\
GP-Init + Refine (greedy) & 25 & \textbf{0.86} (0.06) & 4.6 (2.7) & \textbf{0.58} (0.16) \\
\midrule \midrule
\textit{Upper-bound performance} \\
GeoReg (native) & 400 & 0.87 (0.05) & 6.3 (4.0) & 0.54 (0.21) \\
GP-Init + Refine (greedy) & 400 & 0.87 (0.05) & 3.4 (2.1) & 0.67 (0.17) \\
\bottomrule
\end{tabular}
}
\end{wraptable}

In this work, we have presented GeoPose, a population-trained framework for rapid CTA-to-DSA registration without patient-specific network training or explicit inter-volume preregistration. GeoPose combines canonical-frame pose estimation with projection-space calibration and transform composition to transfer pose predictions directly into the native coordinate frame of an unseen CTA. A population-trained refinement network and low-budget optimization using GeoReg subsequently improve this estimate, enabling registration within approximately two seconds. This rapid CTA-to-DSA registration makes pre-procedural 3D vascular anatomy available in the intraoperative coordinate system. Projecting the anatomy into the acquired C-arm views can support vessel localization and navigation, preserve anatomical context between angiographic acquisitions, and potentially reduce repeated contrast injections.

\begin{figure*}[t]
     \centering
     \includegraphics[width=\textwidth, trim=0.2cm 0.4cm 0 0]{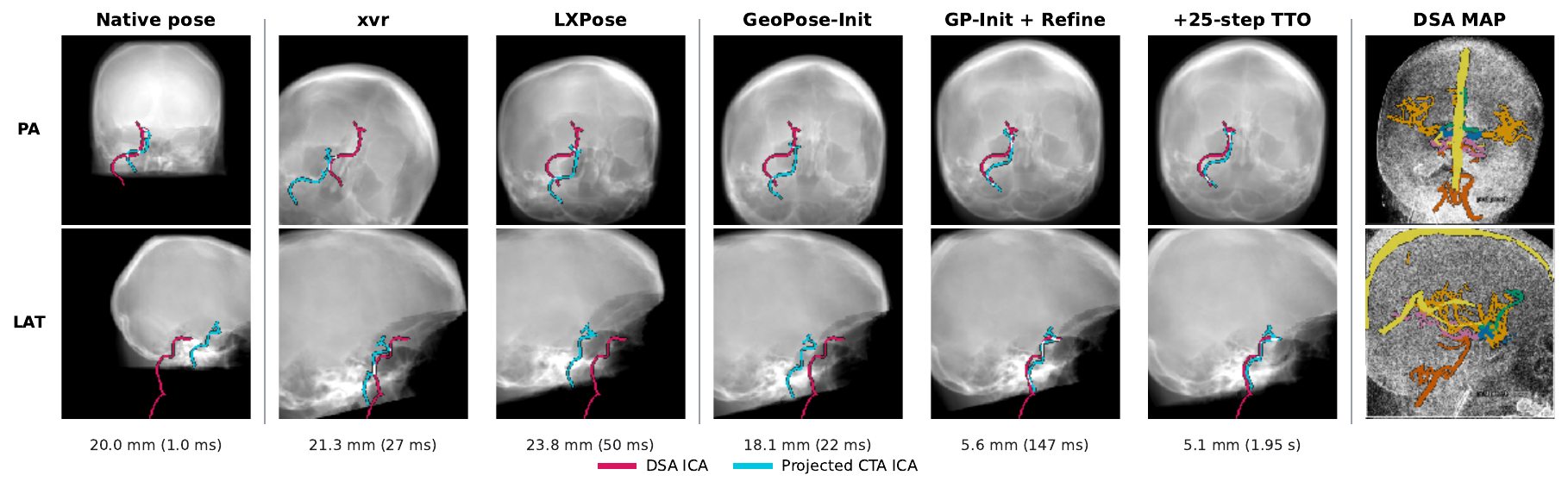}
     \caption{Qualitative progression of CTA-to-DSA registration for the median-performing test subject in terms of mPCD. The first six columns show CTA digitally reconstructed radiographs (DRRs) at the native, xvr, LXPose (net1 + net2), GeoPose-Init, GP-Init + Refine (greedy), and 25-step test-time optimization~(TTO) poses. Vertical rules separate the native pose, baseline initializers, GeoPose-based methods, and final DSA visualization. In the first six columns, DSA-derived internal carotid artery centerlines are shown in magenta and projected CTA centerlines in cyan. The rightmost column projects the CTA-derived TopBrain vascular segmentation~\cite{yang2026topbrain} onto the corresponding DSA maximum-intensity projection using the final 25-step pose. Distinct colors separate major vascular structures, illustrating how the correspondence transfers anatomical information from CTA to DSA. Values below the first six columns report the mean mPCD across the displayed PA and LAT views, followed by the cumulative runtime.}
     \label{fig:qualitative_registration}
\end{figure*}

The results show that the largest improvement in instantaneous pose estimation is obtained through GeoPose-Refine. Although GeoPose-Init performs similarly to LXPose, a single refinement step reduces the average mPCD from 15.7 to 6.9~mm and increases clDice from 0.24 to 0.39 (Table~\ref{tab:instant_pose}). Greedy refinement improves these values to 5.8~mm and 0.45, substantially outperforming LXPose. This improvement is particularly pronounced for the LAT view, for which refinement reduces mPCD from 22.0 to 7.0~mm. In addition to the greater depth ambiguity of a lateral projection, small residual pose deviations in this view can produce a relatively large separation between the projected carotid arteries, and therefore a larger change in mPCD than a comparable deviation in the PA view. GeoPose-Refine is trained using per-axis perturbation distributions calibrated to the residual errors of GeoPose-Init, allowing it to operate directly on the local error regime produced by the initial network. The progression in Fig.~\ref{fig:qualitative_registration} illustrates this behavior: greedy refinement provides most of the visible improvement in vascular overlap, while the subsequent 25-step optimization makes a smaller final correction.

The component analysis in Table~\ref{tab:geopose_ablation} further illustrates the different roles of the initialization and refinement objectives. Adding mPD supervision to greedy refinement reduces mPCD from 8.0 to 5.8~mm and increases clDice from 0.38 to 0.45 compared with Dice-only training. Conversely, the cumulative GeoPose-Init ablation does not show a monotonic improvement across all registration metrics. GeoPose-Init predicts pose from a single projection in the learned canonical frame and is consequently bound by the accuracy of the training-time canonicalization and by how closely the anatomy of an unseen patient agrees with this canonical representation. A downstream refinement model additionally observes a rendering of the patient CTA and can therefore correct residual patient-specific discrepancies that are not available to the initial network. The attainable clDice is similarly limited by differences between the independently obtained CTA and DSA vessel segmentations, which need not coincide perfectly even under accurate registration. Appearance augmentation, including plasma transformations, was additionally important in preliminary experiments for transferring from synthetic DRRs to DSA-derived maximum-intensity projections, although its individual contribution was not isolated in the present ablation.

A central advantage of GeoPose is that these results are obtained without patient-specific network preparation. xvr requires test-time adaptation to the target volume, whereas the original LXPose formulation trains a separate model for each target volume. GeoPose instead retains fixed network weights and uses a single synthetic projection to estimate the frame transfer between the canonical training population and a native CTA. The refinement strategy is similar to LXPose, but preliminary experiments showed that independently encoding the two projections with a shared one-channel encoder and fusing their representations late converged more reliably than direct two-channel input or multi-stage fusion. Initial experiments with joint end-to-end training of GeoPose-Init and GeoPose-Refine provided only limited additional improvement. The refiner is currently trained using isolated pose perturbations and does not explicitly observe the compound errors introduced by canonical-frame prediction and projection-space calibration. Incorporating the complete calibration pathway into training may allow the refiner to learn this error distribution directly, albeit with increased memory requirements.

Initialization and optimization should be regarded as complementary components. After 25 NAdam iterations, native-initialized GeoReg obtains an mPCD of 14.6~mm and a clDice of 0.15, whereas GeoPose with greedy refinement obtains 4.6~mm and 0.58, respectively (Table~\ref{tab:tto_25}). Using only one refinement step before optimization produces nearly identical performance, indicating that greedy refinement is most beneficial as an optimization-free operating point. Moreover, after 400 iterations, native-initialized GeoReg and GeoPose both reach an NCC of 0.87, while GeoPose retains a lower mPCD of 3.4 versus 6.3~mm and a higher clDice of 0.67 versus 0.54. NCC-based convergence therefore does not necessarily correspond to the same anatomical pose, and improved initialization can affect both convergence speed and the final geometric alignment. Greedy refinement accounts for most of the network-only runtime, but still requires only 147~ms for a biplanar pair and is therefore effectively instantaneous.

It should be noted that pose estimation remains sensitive to the observability of individual transformation components. Translation along the source--detector direction (Fig.~\ref{fig:tto}(d), $\Delta y$) and coupled out-of-plane rotation (Fig.~\ref{fig:tto}(h), $\Delta\theta_z$) are more difficult to infer from a single projection, consistent with observations for DiffPose and xvr \cite{gopalakrishnan2024intraoperative,gopalakrishnan2025rapid}. Different rotational parameterizations had little effect in our experiments, likely because GeoPose regresses relatively small deviations from a nearby view-dependent isopose rather than rotations over the full pose space. Out-of-distribution results additionally demonstrate a remaining domain gap. Training on TopCoW and evaluating on the reserved ISLES'24 test set increases mPCD from 5.8 to 8.5~mm and reduces clDice from 0.45 to 0.28 (Table~\ref{tab:instant_pose}), with a further decrease on the broader ISLES'24 cohort. Differences in spatial resolution, field of view, and vascular coverage are likely contributors, as TopCoW emphasizes the Circle of Willis, whereas the matched ISLES'24 and DSA data contain broader and more variable cranial coverage. Finally, calibration is estimated from a single PA rendering, such that residual calibration errors may propagate into both view predictions. Future work may investigate multi-view calibration to improve robustness to these sources of error.

Reliable recovery of biplanar acquisition geometry can provide immediate value during EVT. GeoPose registers DSA data directly to CTA renderings without relying on vascular segmentations or explicit vessel correspondences. The estimated poses allow CTA-derived 3D segmentations and treatment targets to be projected onto the acquired DSA views, where their anatomical labels distinguish individual vascular structures (Fig.~\ref{fig:qualitative_registration}). The same geometry can be used to reconstruct 3D vasculature from paired DSA sequences without a separate rotational acquisition~\cite{frisken2025spatiotemporally}. Future work may incorporate contrast dynamics across the registered views to reconstruct three-dimensional contrast transport and estimate patient-specific blood flow~\cite{de2023spatio,de2024accelerating}. In conclusion, GeoPose delivers accurate native-frame registration in approximately two seconds without patient-specific training or explicit preregistration, making pre-procedural 3D anatomy directly available in the DSA acquisition geometry for image-based guidance and biplanar 3D reconstruction.

\section*{Acknowledgments}
This study was supported by ZonMw Rubicon under grant no.~04520252520006. Rudolf L. M. van Herten would like to acknowledge Vivek Gopalakrishnan’s thoughtful feedback during the final revision of this manuscript.

\bibliographystyle{unsrt}
\bibliography{references}

\end{document}